\documentclass[sigconf,natbib=true,anonymous=false]{acmart}
\AtBeginDocument{%
  }

\setcopyright{acmlicensed}
\copyrightyear{2025}
\acmYear{2025}
\acmDOI{XXXXXXX.XXXXXXX}
\acmConference[XXX'25]{XXX}{XXXX XX--XX,
  2025}{XXXX, XX}
\acmISBN{978-1-4503-XXXX-X/2025/06}

\usepackage{graphicx}
\usepackage{booktabs}
\usepackage{multirow,array}
\usepackage[table]{xcolor}
\usepackage{enumitem}
\usepackage{float}
\usepackage{amsmath}
\usepackage{amsfonts}
\usepackage{algorithm}
\usepackage{algorithmicx}
\usepackage{algpseudocode}
\usepackage{tabularx}
\usepackage{color, xcolor}
\usepackage{caption}

\begin{document}

\title{Listwise Preference Optimization with Element-wise Confusions for Aspect Sentiment Quad Prediction}

\author{Wenna Lai}
\affiliation{%
  \institution{\centering\parbox[t]{5.6cm}{The Hong Kong Polytechnic University}}
  \city{Hong Kong SAR}
  \country{China}
}
\email{winnelai05@gmail.com}

\author{Haoran Xie}
\authornote{Corresponding author}
\affiliation{%
  \institution{Lingnan University}
  \city{Hong Kong SAR}
  \country{China}
  }
\email{hrxie@ln.edu.hk}

\author{Guandong Xu}
\affiliation{%
  \institution{University of Technology Sydney}
  \city{Sydney}
  \country{Australia}
}
\email{guandong.xu@uts.edu.au}
\affiliation{%
  \institution{\centering\parbox[t]{5.6cm}{The Education University of Hong Kong}}
  \city{Hong Kong SAR}
  \country{China}
}
\email{gdxu@eduhk.hk}

\author{Qing Li}
\affiliation{%
   \institution{\centering\parbox[t]{5.6cm}{The Hong Kong Polytechnic University}}
  \city{Hong Kong SAR}
  \country{China}
  }
\email{qing-prof.li@polyu.edu.hk}

\author{S. Joe Qin}
\affiliation{%
   \institution{Lingnan University}
  \city{Hong Kong SAR}
  \country{China}
  }
\email{joeqin@ln.edu.hk}

\renewcommand{\shortauthors}{Lai et al.}

\begin{abstract}
Aspect sentiment quad prediction (ASQP) is inherently challenging to predict a structured quadruple with four core sentiment elements, including aspect term ($a$), aspect category ($c$), opinion term ($o$), and sentiment polarity ($s$). Prior methods relying on marker-based prediction struggle with modeling the intricate relationships among elements and experience sharp performance declines when predicting higher-order elements (e.g., $c$ and $s$) under standard supervised fine-tuning. To address these limitations, we employ reasoning-based generation to output both the quadruple and a natural language rationale under element prefixes within a unified template, encouraging explicit relational reasoning and interpretability. To further enhance element-wise alignment, we introduce a listwise preference optimization framework for improving structural validity and relational coherence. Specifically, we generate element-wise confusable candidates via syntactic and semantic proximity, then train the model with listwise objectives to prefer the gold candidates over closely competing alternatives. Extensive experiments on four benchmark datasets demonstrate that our framework effectively improves quadruple prediction accuracy and explanation consistency. Our code is released at https://anonymous.4open.science/r/E4L.
\end{abstract}



\maketitle


\section{Introduction}

\begin{figure}[ht]
  \centering
  \includegraphics[width=\columnwidth]{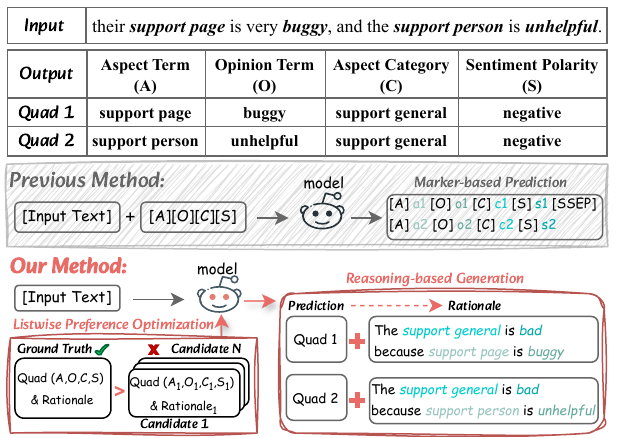}
  \caption{An example for the ASQP task. Previous methods remain maker-based prediction, where the extracted quads may be structurally valid but semantically incoherent, reflected as incorrect relationships in rationales.}
  \label{fig:example}
  \vspace{-3mm}
\end{figure}

Aspect-based sentiment analysis (ABSA) has become a prominent research focus within natural language processing (NLP), aiming to extract fine-grained sentiment information at the aspect level \cite{tkde/SchoutenF16, tkde/ZhangLDBL23, lai2025llmsteamup}. This field has drawn considerable interest because of its utility in applications such as opinion mining and customer feedback analysis \cite{rvisa, naacl/ZhangDLPB24, lai2024mtisa}. Among various ABSA tasks, aspect sentiment quad prediction (ASQP) is particularly challenging to extract a structured quadruple comprising the \emph{aspect term (a)}, \emph{aspect category (c)}, \emph{opinion term (o)}, and \emph{sentiment polarity (s)} \cite{Peng_Xu_Bing_Huang_Lu_Si_2020, emnlp/ZhangD0YBL21}. As illustrated in Figure \ref{fig:example}, a single sentence often contains multiple quads with different opinions and relationships, e.g., ``\emph{their support page is very buggy, and the support person is unhelpful.}'' The corresponding quads include (\emph{support page, buggy, support general, negative)} and \emph{(support person, unhelpful, support general, negative)}, each representing \emph{(a, o, c, s)} in order. This task is inherently complex because the model requires not only identifying all four elements in multiple quads but also capturing the dependencies among sentiment elements.

Recent research has highlighted that the performance of existing ASQP methods is often constrained by two primary challenges: limited annotated data \cite{acl/MvP, self-scorer} and restricted adaptability to unseen targets \cite{acl/bsvp}. Even with large language models (LLMs) demonstrating impressive generative and reasoning capabilities across many NLP tasks, their performance on ASQP remains modest, particularly as the complexity of the output structure increases \cite{Xu2023TheLO, naacl/ZhangDLPB24}. To mitigate these issues, most previous studies employed data augmentation strategies. For example, Gou et al. \cite{acl/MvP} used multi-view prompting to vary element orders. Zhang et al. \cite{self-scorer} developed a pseudo-label scorer trained on additional comparison datasets to assist label selection during inference. Bai et al. \cite{acl/bsvp} curated more balanced few-shot datasets to improve adaptation to unseen aspect categories. Instead of augmenting data instances, Jian et al. \cite{SimRP} retrieved similar demonstrations to enrich input prompts. Despite these advances, existing ASQP methods predominantly adopt marker-based prediction via supervised fine-tuning (SFT), which tends to treat each element in isolation, as shown in Figure \ref{fig:example}. This often leads to mismatches and inconsistencies among sentiment elements, and a lack of reasoning for the predicted quads. Consequently, the extracted quads may be structurally valid but semantically incoherent, limiting their reliability and interpretability. 

To overcome these limitations, we adopt reasoning-based generation for ASQP within a unified natural language template, linking all sentiment elements with an explicit rationale target guided by element prefixes. This structure explicitly binds rationales to each sentiment quad, enabling reasoning through both semantic and relational dimensions to promote consistency in explanations.
Motivated by preference modeling from Direct Preference Optimization (DPO), we posit that generative ASQP naturally entails preferences over alternative outputs with respect to \emph{structural validity} and \emph{relational coherence}. \textit{\textbf{Structural validity}} ensures that the predicted output is complete and conforms to the template. \textit{\textbf{Relational coherence}} assesses whether the predicted elements are accurate to reflect a coherent semantic and relational chain. Standard SFT objectives do not encode these preferences, and pairwise preference optimization (e.g., DPO) is insufficient to capture global rankings over a set of closely confusable candidates. 

In order to better align model training with these preferences, we introduce a listwise preference optimization framework for ASQP that leverages element-wise confusable candidates generated via syntactic and semantic proximity. For each input, we construct a gold output and a set of hard negatives by minimally altering one or more elements and correspondingly updating the rationale within the unified reasoning-based template. This setup exposes fine-grained confusions that challenge the model understanding of element dependencies and explanation consistency. The model is then trained with a listwise objective to prefer the gold candidate over all confusable alternatives, amplifying the probability of the correct quadruple and rationale.


In summary, our contributions are as follows:
\begin{itemize}[label={}, left=1em]
    \item 1) To the best of our knowledge, this work is the first to integrate reasoning-based generation with preference optimization for ASQP, effectively guiding structured prediction and reasoning.
    \item 2) We develop \underline{\textbf{E}}lement-wise Confusable Candidates \underline{\textbf{for}} \underline{\textbf{L}}istwise Preference Optimization (\textbf{E4L}), a framework that explicitly models inter-element relationships to enhance both structural validity and relational coherence.
    \item 3) Extensive experiments on four benchmark datasets demonstrate the effectiveness of our framework in improving quad prediction performance and explanation consistency.
\end{itemize}

\section{Preliminary}
Given an ASQP dataset, the model is trained to predict a set of four sentiment elements, including \emph{aspect term} ($a$), \emph{opinion term} ($o$), \emph{aspect category} ($c$), and \emph{sentiment polarity} ($s$). The predicted quadruple set is denoted by $Q = \{(a,o,c,s)\}$. Following the prior work \cite{emnlp/ZhangD0YBL21}, we map sentiment elements to semantic expressions. For instance, the sentiment polarity ``\textit{positive}'' is expressed as ``\textit{great}'' in the target sequence, and the ``\textit{NULL}'' label is expressed as ``\textit{it}''. Consequently, the prediction target becomes $y = Q = \{(m_a,m_o,m_c,m_s)\}$, where $m$ denotes the mapping function. During inference, the target $(m_a,m_o,m_c,m_s)$ is converted back to the original elements $(a,o,c,s)$ for evaluation. In general, $a$ and $o$ are explicit text spans from the input sentence $x$, except for implicit cases where they are predicted as ``\textit{NULL}''.
\section{Methodology}
\begin{figure*}[t]
  \includegraphics[width=\linewidth]{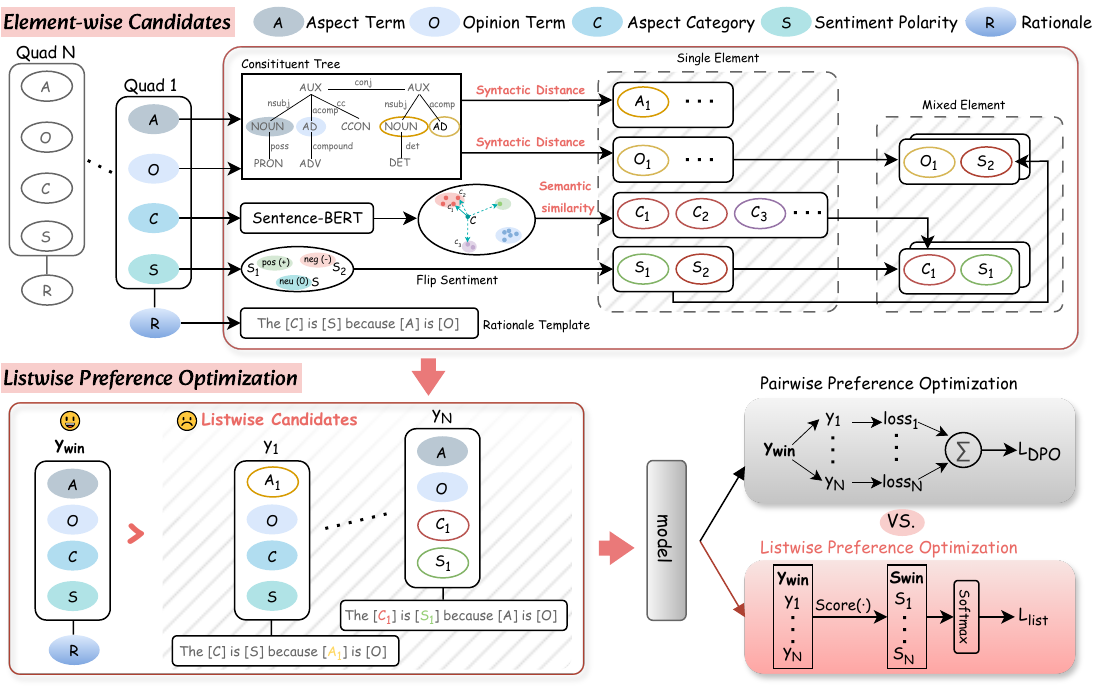}
  \caption {An overview of our proposed method (E4L). We first curate element-wise candidates via syntactic and semantic proximity, then compose them into listwise candidates for preference optimization.}
  \label{fig:method}
\end{figure*}

\paragraph{\textbf{Motivation.}}
Previous methods are predominantly marker-based, where distinct markers $T_a, T_o, T_c, T_s$ are assigned to each element to differentiate them during generation. The prediction target becomes
\begin{equation}
    y = [T_a] a [T_o] o [T_c] c [T_s] s [\mathrm{SSEP}] \ldots
\end{equation}
This approach tends to treat each element in isolation, adhering to each element marker but without paying attention to inter-element dependencies and reasoning. As a result, the predicted quads may be structurally valid but lack relational coherence. 

To address these limitations, we develop a reasoning-based natural language template guided by element prefixes to link each quad $Q$ with an explicit rationale target $r = R(Q)$.
We adopt a paraphrase template from \cite{emnlp/ZhangD0YBL21} to express the rationale, as it entails all elements and captures the necessary causal relationships. Then the prediction objective becomes:
\begin{equation}
    \label{prediction obj}
    y = p(Q, r \mid prompt(x))
\end{equation}
This encourages the model to learn the correct elements alongside their relationships reasoning, enhancing both consistency and interpretability.

However, such a setup with SFT-only still faces the challenge of selecting among multiple plausible element-wise candidates. We posit that generative ASQP inherently involves preferences over the set of alternatives concerning \emph{structural validity} and \emph{relational coherence}. To better align model training with these preferences, we introduce a listwise preference optimization framework for ASQP, as shown in Figure \ref{fig:method}. We construct element-wise confusable candidates through syntactic and semantic proximity, and train the model to prefer the gold outputs over these alternatives with a listwise objective. The following sections will detail the framework.

\subsection{Element-wise Candidates}
\label{element-wise}
To prepare preference data for optimization, we generate a comprehensive set of confusable candidates, ensuring coverage across all constituent elements. As syntax and semantics are both essential in natural language understanding, we incorporate syntactic distance and semantic similarity when constructing element-wise confusions.

\paragraph{\textbf{Syntatic Distance.}}
In NLP, syntactic structure is often represented by constituent trees ($CT$), which capture the hierarchical organization of sentence components. Tools such as \textit{spaCy} can parse these trees efficiently. For lower-order elements like aspect terms $a$ and opinion terms $o$, which typically appear as spans in the sentence, syntactic proximity is effective for selecting confusable candidates, especially in sentences containing multiple quads.

Given an input sentence $x$, we obtain its constituent tree via:
\begin{equation}
    CT(x) = \mathrm{Parser}(x).
\end{equation}

Using this parse, we search the candidate spans $C_s$ from $CT$ that match the part-of-speech (POS) pattern of the gold element $e$. Then, we measure the syntactic distance between each span $s_i \in C_s$ and the gold element $e$ to identify plausible confusing candidates $C$:
\begin{equation}
    C(e) =  \mathrm{Top}\text{-}k \space \arg\min_{s_i \in C_s}\mathrm{syn}(e,s_i),
\end{equation}
where $\mathrm{syn}(e,s_i) = \|idx(e), idx(s_i)\|$ denotes tree distance and $idx(\cdot)$ is the corresponding index in $CT(x)$. Candidates with minimal syntactic distance or similar parse structures are more likely to be confusable, thereby challenging the model to discriminate between syntactically similar but incorrect options.

\paragraph{\textbf{Semantic Similairty.}}
While syntactic distance captures structural similarity, semantic similarity ensures contextual relevance. Considering the aspect category $c$ is often not explicitly present in the sentence, it is pre-defined within a fine-grained category list $L_c$ and requires contextual reasoning from preceding elements to determine the belonging category. Semantic similarity can therefore guide the selection of confusable categories from a given list. 

We use a pretrained encoder (e.g., Sentence-BERT \cite{Sentence-BERT}) to obtain dense representations of the targeted element $e$, and compute the semantic embedding:

\begin{equation}
    \mathbf{Emb}(e) = \mathrm{Sentence-BERT}(e).
\end{equation}

Then, compute cosine similarity between the target category $c$ and each candidate $L_c^{(i)} \in L_c$:

\begin{equation}
\mathrm{sim}(c, L_c^{(i)}) = \frac{\mathbf{Emb}(c) \cdot \mathbf{Emb}(L_c^{(i)})}{\|\mathbf{Emb}(c)\| \; \|\mathbf{Emb}(L_c^{(i)})\|},
\label{similarity}
\end{equation}
and select candidates $C$ as:
\begin{equation}
    C(c) = \mathrm{Top-}k \space \arg\max_{{L_c}_i \in L_c} \space \mathrm{sim}(c, {L_c}_i).
\end{equation}

\noindent For sentiment polarity $s$, we generate single-element confusion by flipping $s$ to other labels sampled from the label list. The rationale $r$ that references all sentiment elements is synchronized accordingly. To increase the candidate complexity for more robust model learning, we also develop mixed-element confusions from single-element candidates. See Algorithm \ref{element-wise algo} for the detailed construction process.

\begin{algorithm}[ht]
\small
\caption{Element-wise Candidates}
\raggedright
\label{element-wise algo}
\textbf{Given:} ASQP dataset $D = \{(x_i, y_i)\}^N$, where $y_i = \{(a, o,c,s)^m\}$; aspect category list $L_c$; sentiment polarity list $L_s$; syntactic distance function $\mathrm{syn}(\cdot)$; semantic similarity function $\mathrm{sim}(\cdot)$; constituent tree parser $\mathrm{Parser}(\cdot)$.\\
\textbf{Output:} Element-wise candidates $C = \{C^j_i\}$ for each quad.
\begin{algorithmic}[1]
\For{$(x_i, y_i)$ in $D$} 
\For{$(a,o,c,s)^j$ in $y_i$}
\State Initialize element-wise candidates $C^j_i = \emptyset$
\State $CT(x_i) = \mathrm{Parser}(x_i)$ 
\State \textbf{// Single-element candidates}
\For{ $e$ in $(a, o)$}
\State $C_s = \mathrm{Search}(CT(x_i), POS(e))$ 
\State $C(e) = \mathrm{Top-}k \arg\min_{s_l \in C_s}\mathrm{syn}(e,s_l)$
\State $C^j_i \gets C^j_i \cup \{C(e)\}$ \Comment{get $C(a)$ and $ C(o)$}
\EndFor
\State $C(c) = \mathrm{Top-}k \arg\max_{c_l \in L_c}\mathrm{sim}(c,c_l)$
\State $C(s) = \{ s_l \in L_s \mid s_l \neq s\}$
\State $C^j_i \gets C^j_i \cup \{C(c)\}$ \Comment{get $C(c)$}
\State $C^j_i \gets C^j_i \cup \{C(s)\}$ \Comment{get $C(s)$}
\State \textbf{// Mixed-element candidates}
\State $C(o,s) = \emptyset$
\For{ $o'$ in $C(o)$}
\State $s' = \mathrm{Top-}k \arg\max_{s_c \in C(s)}\mathrm{sim}(o',s_c) $
\State $C(o,s) \cup {(o', s')}$  \Comment{get $C(o,s)$}
\EndFor
\State $C(c,s) = \emptyset$
\For{ $c'$ in $C(c)$}
\State $s' \gets \text{Sample}(C(s))$
\State $C(o,s) \cup {(c', s')}$ \Comment{get $C(c,s)$}
\EndFor
\State $C^j_i \gets C^j_i \cup \{C(o,s)\}$  
\State $C^j_i \gets C^j_i \cup \{C(c,s)\}$ 
\EndFor
\EndFor
\end{algorithmic}
\end{algorithm}

\subsection{Listwise Preference Optimization}
With element-wise confusable candidates $C$ from Section \ref{element-wise}, we compose listwise candidates at the quad level as targets. For each input $x$, we curate a set of hard negatives $y_{l} = {y_1,...y_N}$ by minimally altering one or more elements of the gold output $y_w = (Q_w, r_w)$ from $C$, and synchronize the rationale $r$ accordingly. This setup exposes fine-grained confusions that challenge the model understanding of inter-element dependencies and explanation consistency.

Given an input $x$, the objective of supervised fine-tuning (SFT) follows Equation (\ref{prediction obj}), where we train a base language model by:
\begin{equation}
    \mathcal{L}_{\mathrm{CE}} = - \mathbb{E}_{(x, y_w)} \left[ \log \pi_\theta (y_w \mid x) \right],
\end{equation}

For DPO, given a reference policy $\pi_\mathrm{ref}$, the objective is to train a target policy $\pi_\theta$ that aligns with human preferences while remaining close to $\pi_\mathrm{ref}$, where the pairwise preference is that $y_w \succ y_i$ for each $ y_i \in y_{l}$, and the loss $\mathcal{L}_{\mathrm{DPO}}(\theta)$ is:

\begin{equation}
\begin{aligned}
     - \mathbb{E}\left[ \log \sigma \left( \beta \log \frac{\pi_\theta(y_w | x)}{\pi_\mathrm{ref}(y_w| x)} - \beta \log \frac{\pi_\theta(y_l | x)}{\pi_\mathrm{ref}(y_l | x)} \right) \right]
\end{aligned}
\end{equation}

However, DPO is pairwise and does not exploit global ranking among multiple confusable candidates. To better capture the inherent preferences regarding structural validity and relational coherence, a listwise objective is introduced to maximize the probability of the gold output  $y_w$ over the entire candidate set $S$, leveraging the full ranking information.

Let $S=\{y_w\} \cup  y_l$ with a size of $N+1$, we assign the log-ratio of policy as the implicit reward score $s_i$ for each candidate $y_i \in S$:
\begin{equation}
s_i = \beta \log \frac{\pi_\theta(y_i \mid x)}{ \pi_\mathrm{ref}(y_i \mid x)}.
\end{equation}

The distribution over candidates within the candidate set $S$ is obtained via a softmax:
\begin{equation}
P_\theta(y_i \mid x) = \frac{\exp(s_i)}{\sum_{j=1}^{N+1} \exp(s_j)},
\label{listwise_distribution}
\end{equation}

To optimize the policy $\pi_\theta$ for maximizing $P_\theta(y_w|x)$, the listwise objective is defined as follows:
\begin{equation}
    \mathcal{L}_{\mathrm{list}}(\theta) = -\mathbb{E}\left[\sum_{i=1}^N p(y_i|x)\cdot logP_\theta(y_i|x)\right],
\end{equation}
where $p(y_i|x)$ is the probability distribution that determines listwise human preferences among all candidates. For ensuring prediction accuracy, we use a one-hot target distribution that places all mass on the gold output $y_w$: 
\begin{equation}
    p(y_i|x) = \left\{ \begin{aligned}
        1,& \quad y_i = y_w,\\
        0,& \quad y_i \in \{y_l\}
    \end{aligned}\right.
\end{equation}
Then, the listwise loss is:
\begin{equation}
    \mathcal{L}_{\mathrm{list}}(\theta) = -p(y_w\mid x) \cdot logP_\theta(y_w\mid x).
\end{equation}

To further stabilize preference optimization and maintain structural validity, we combine the listwise objective loss with a hybrid cross-entropy term used in SFT:
\begin{equation}
    \mathcal{L}(\theta) = (1-\lambda)\mathcal{L}_{list}(\theta) + \lambda \mathcal{L}_{\mathrm{CE}},
\end{equation}
where $\lambda \in [0,1]$ controls the trade-off between preference alignment and supervised generation.

\section{Experiments}
\subsection{Datasets and Metrics} The evaluation datasets contain four publicly available ASQP datasets, including \textsc{ASQP-Rest15}, \textsc{ASQP-Rest16}, \textsc{ACOS-Laptop}, and \textsc{ACOS-Rest}, which are derived from the SemEval Challenges \cite{semeval/PontikiGPMA15,semeval/PontikiGPAMAAZQ16} and the Amazon platform. Among them, \textsc{ACOS-Laptop} and \textsc{ACOS-Rest} are developed for aspect-category-opinion-sentiment (ACOS) quadruple extraction \cite{acl/CaiXY20}, which contains a greater number of implicit aspects and opinions in expressions. Dataset statistics are provided in Table \ref{tab:asqp-stat}. A prediction is counted as correct only under an exact match of all four elements with the gold quadruple. The evaluation metric uses precision (Pre), recall (Rec), and the F1 score. 

\subsection{Baselines} We compare our methods against several state-of-the-art approaches. These baselines include non-generative, generative, and collaborative methods.

\paragraph{Non-generative methods}
\begin{itemize}
    \item \textbf{EXTRACT-CLASSIFY} \cite{acl/CaiXY20}: a pipeline approach that first extracts candidate elements and then performs classification.
    \item \textbf{OTCL} \cite{OTCL_li}: introduces Opinion-Tree-guided contrastive learning to enhance text representations. 
    \item \textbf{IVLS} \cite{IVLS_NieFZL24}: formulates the task with a pointer-based, non-autoregressive framework.
\end{itemize}
\paragraph{Generative methods}
\begin{itemize}
    \item \textbf{GAS} \cite{acl/Zhang0DBL20}: the first approach to model the ABSA task as a text generation problem.
    \item \textbf{PARAPHRASE} \cite{emnlp/ZhangD0YBL21}: linearizes structured quads into natural-language sequences for generation.
    \item \textbf{SEQ2PATH} \cite{acl/MaoSYZC22}: generates tuples by producing paths on a tree structure.
    \item \textbf{DLO/ILO} \cite{acl/HuBWZZGZH23}: selects effective element orders and leverages multiple templates for data augmentation. 
    \item \textbf{GENDA} \cite{starsem/WangJMLO23}: augments parallel training data via generation-based methods. 
    \item \textbf{MUL} \cite{acl/HuBWZZGZH23}: controls token-level generation by accounting for model uncertainty.
    \item \textbf{SimRP} \cite{SimRP}: uses syntactic and semantic retrieval to improve in-context demonstration selection. 
\end{itemize}
\paragraph{Collaborative methods}
\begin{itemize}
    \item \textbf{SCRAP} \cite{self-consistency}: incorporates self-consistency to help models internalize LLM-provided rationales.
    \item \textbf{GAS+Scorer\&ReRank(AI)} \cite{self-scorer}: employs a trained scorer built on AI-generated data to assist GAS \cite{acl/Zhang0DBL20} in label selection.
    \item \textbf{MUL+Scorer\&ReRank(AI)} \cite{self-scorer}: applies the same scorer to MUL \cite{acl/HuBWZZGZH23} for label selection.
\end{itemize}

\begin{table}[h]
\centering
\caption{
Statistics of four ASQP datasets \cite{emnlp/ZhangD0YBL21, acl/CaiXY20}. 
\#S and \#Q represent the number of sentences and quads. 
}
\begin{tabular}{l cc c cc c cc} 
\toprule
\multirow{2}*{{Datasets}} & \multicolumn{2}{c}{{Train}} && \multicolumn{2}{c}{{Dev}} && \multicolumn{2}{c}{{Test}} \\
\cmidrule(r){2-3} \cmidrule(r){5-6} \cmidrule(r){8-9}
& \#S & \#Q && \#S & \#Q  && \#S & \#Q\\
\midrule
\texttt{ASQP-Rest15} & 834 & 1354 && 209 & 347 && 537 & 795 \\
\texttt{ASQP-Rest16} & 1264 & 1989 && 316 & 507 && 544 & 799\\
\texttt{ACOS-Laptop} & 2934 & 4172 && 326 & 440 && 816 & 1161\\
\texttt{ACOS-Rest}   & 1530 & 2484 && 171 & 261 && 583 & 916\\
\bottomrule
\end{tabular}
\label{tab:asqp-stat}
\end{table}
\subsection{Implementation Details} We use Flan-T5-large \cite{flant5} from HuggingFace Transformers as the backbone model. Considering the computational cost, we set the number of listwise candidates to $N =6$, ensuring each element contributes at least one confusable candidate. In SFT, we train for $20$ epochs with a batch size of $64$ and a learning rate of $1 \times 10^{-4}$, using the AdamW optimizer. For preference optimization, both pairwise and listwise are run for $3$ epochs with a batch size of $64$ and a learning rate of $1 \times 10^{-6}$. We use the AdamW optimizer, and a warmup ratio of $0.1$. A grid search is applied for selecting $\lambda$. All experiments are conducted on $4\times80$GB Nvidia H800 GPUs.

\begin{table*}[h]
\centering
\caption{
Main experimental results on four benchmark datasets (\%), where \textsc{ACOS} datasets are more challenging with a greater number of implicit aspects and opinions in expressions. Bold values highlighted in \textit{pink} indicate the overall best F1 score, reflecting the balance between precision (Pre) and recall (Rec). Values highlighted in \textit{blue} denote the second-best F1 score.}
\setlength\tabcolsep{2.8pt}
\begin{tabular}{l c ccc c ccc c ccc c ccc} 
\toprule
\multirow{2}*{{Methods}} && \multicolumn{3}{c}{ASQP-Rest15} && \multicolumn{3}{c}{ASQP-Rest16}  && \multicolumn{3}{c}{ACOS-Laptop} && \multicolumn{3}{c}{ACOS-Rest} \\
\cmidrule(r){3-5}  \cmidrule(r){7-9} \cmidrule(r){11-13} \cmidrule(r){15-17}
&& \texttt{Pre} & \texttt{Rec} & \texttt{F1} && \texttt{Pre} & \texttt{Rec} & \texttt{F1} && \texttt{Pre} & \texttt{Rec} & \texttt{F1} && \texttt{Pre} & \texttt{Rec} & \texttt{F1} \\
\midrule
 \bf\emph{- Training-free baselines}&&&&&& \\
\textsc{ChatGPT} (few-shot) \cite{Xu2023TheLO} & & 29.66 & 37.86 & 33.26 & & 36.09 & 46.93 & 40.81 && 21.72 & 27.65 & 24.33 & & 38.39 & 46.40 & 42.02 \\
Llama-3.1-70B (10-shot) \cite{naacl/ZhangDLPB24}&& -& -&32.84&&-&-&37.10&& -& -&-&&-&-&-\\
GPT-4o (10-shot) \cite{naacl/ZhangDLPB24}&& -& -&37.08&&-&-&45.00&& -& -&-&&-&-&-\\
\midrule
\bf\emph{- State-of-the-art baselines }&&&&&& \\

\textsc{Extract-Classify} \cite{acl/CaiXY20} && 35.64 & 37.25 & 36.42 && 38.40 & 50.93 & 43.77 && 45.56 & 29.28 & 35.80 && 38.54 & 52.96 & 44.61 \\
\textsc{Paraphrase} \cite{emnlp/ZhangD0YBL21} && 46.16 & 47.72 & 46.93 && 56.63 & 59.30 & 57.93 && - & - & - && - & - & - \\
\textsc{GAS} \cite{acl/Zhang0DBL20} && 47.15 & 46.01 & 46.57 && 57.30 & 57.82 & 57.55 && 43.46 & 42.69 & 43.07 && 59.81 & 57.51 & 58.63 \\ 
\textsc{DLO} \cite{emnlp/DLO} && 47.08 & 49.33 & 48.18 && 57.92 & 61.80 & 59.79 && 43.40 & 43.80 & 43.60 && 60.02 & 59.84 & 59.18 \\
\textsc{ILO} \cite{emnlp/DLO} && 47.78 & 50.38 & 49.05 && 57.58 & 61.17 & 59.32 && 44.14 & 44.56 & 44.35 && 58.43 & 58.95 & 58.69 \\
\textsc{Seq2Path} \cite{acl/MaoSYZC22} && - & - & -  && - & - & - && - & - & 42.97 && - & - & 58.41 \\
\textsc{LEGO-ABSA} \cite{coling/GaoFLLLLBY22} && - & - & 45.80 && - & - & 57.70 && - & - & - && - & - & - \\
\textsc{MvP} \cite{acl/MvP} && - & - & 51.04 && - & - & 60.39 && - & - & 43.92 && - & - & \cellcolor{blue!10}61.54 \\
\textsc{GenDA} \cite{starsem/WangJMLO23} && 49.74 & 50.29 & 50.01 && 60.08 & 61.70 & 60.88 && - & - & - && - & - & - \\

MUL \cite{acl/HuBWZZGZH23} && 49.12 & 50.39 & 49.75 && 59.24 & 61.75 & 60.47 && 44.38 & 43.65 & \cellcolor{blue!10}44.01 && 61.22 & 59.87 & 60.53 \\

OTCL \cite{OTCL_li} && 47.86&50.77&49.27 &&58.31&62.02 &60.11&& -& -&-&&-&-&-\\
IVLS \cite{IVLS_NieFZL24} &&54.46&48.53&51.28 &&62.69&59.75&61.04&&	53.47&33.45&43.71&&60.46&51.14&55.25\\

SimRP \cite{SimRP} && 53.12&53.50& \cellcolor{blue!10}53.30 && 62.74&64.12& \cellcolor{blue!10}63.42&& -& -&-&&-&-&-\\

\midrule
E4L (Ours) && 54.12	&55.35	& \cellcolor{pink!40}\textbf{54.73} && 63.98 & 64.46 & \cellcolor{pink!40}\textbf{64.21} && 45.38 & 46.08 & \cellcolor{pink!40}\textbf{45.73} && 65.46& 	64.37 & \cellcolor{pink!40}\textbf{64.91} \\

\bottomrule
\end{tabular}
\label{tab:main-result}
\end{table*}

\subsection{Main Results}
\subsubsection{Compare to Baselines.} We present the main experimental results compared to baseline methods in Table \ref{tab:main-result}. Our proposed approach consistently outperforms baselines across all four datasets and every evaluation metric, demonstrating its effectiveness in improving structured reasoning for ASQP tasks.
Among the baselines, SimRP \cite{SimRP} leverages syntactic and semantic retrieval to identify similar demonstrations, enriching input prompts to enhance model comprehension. In comparison, our method introduces a more fine-grained, element-level training signal by purposefully constructing element-wise confusing examples, thereby challenging the model to better distinguish between syntactically and semantically similar targets. Moreover, LEGO-ABSA \cite{coling/GaoFLLLLBY22} and MVP \cite{acl/MvP} adopt multi-task learning to improve fine-tuning performance through increased data volume, with MVP further augmenting data instances for each task. Although these methods achieve notable improvements, our approach yields more substantial gains by applying listwise preference optimization beyond standard SFT. Crucially, this improvement requires no extra data source. It is achieved by aligning the overall ranking of the target quadruple with its synchronized rationale among all confusable element-wise candidates, a process that inherently emphasizes structural validity and relational coherence.

\begin{table}[h]
\centering
\caption{
Comparison with collaborative methods on four ASQP datasets with F1 score metric (\%), where the results are obtained from the source papers \cite{self-consistency, self-scorer}. The bold text indicates the best result.}
\setlength\tabcolsep{2.8pt}
\begin{tabular}{lcccc} 
\toprule
\multirow{2}*{{Methods}} & \multicolumn{2}{c}{ASQP} & \multicolumn{2}{c}{ACOS}  \\
\cmidrule(r){2-3}  \cmidrule(r){4-5} 
& \texttt{Rest15} & \texttt{Rest16}&  \texttt{Laptop} & \texttt{Rest} \\
\midrule
SCRAP  &49.93&62.48&-&-\\
GAS+Scorer\&ReRank (AI) & 51.74& 63.51&46.01& 62.47 \\
MUL+Scorer\&ReRank (AI)  &51.97&63.88&\textbf{46.17}&63.63 \\
\midrule
E4L (Ours) & \textbf{54.73} &  \textbf{64.21} & 45.73 & \textbf{64.91} \\
\bottomrule
\end{tabular}

\label{tab:compare collab}
\end{table}

\begin{table}[h]
\centering
\caption{
Ablation study on four ASQP datasets with F1 score metric (\%). The bold text indicates the best result, while the underlined text represents the second-best result.}
\setlength\tabcolsep{2.8pt}
\begin{tabular}{lcccc} 
\toprule
\multirow{2}*{{Methods}} & \multicolumn{2}{c}{ASQP} & \multicolumn{2}{c}{ACOS}  \\
\cmidrule(r){2-3}  \cmidrule(r){4-5} 
& \texttt{Rest15} & \texttt{Rest16}&  \texttt{Laptop} & \texttt{Rest} \\
\midrule
E4L (Ours) & \textbf{54.73} &  64.21 & \textbf{45.73} & \textbf{64.91} \\
\quad-w/o Listwise Loss & \underline{54.49} &\textbf{64.35}& \underline{45.41} &64.79\\
\quad -w/o Pairwise DPO &54.22 & \underline{64.25} &  45.14 &\underline{64.79}\\

\quad -w/o DPO \& Rationale & 53.92&63.29 &44.24&63.26\\

\bottomrule
\end{tabular}
\vspace{0.5em}
\label{tab:ablation study}
\end{table}

\subsubsection{Compare to Collaborative Methods.}
To further illustrate the potential of our method, Table \ref{tab:compare collab} compares its performance with collaborative methods \cite{self-consistency, self-scorer}, which either train a model using data generated by a more advanced LLM or integrate a well-trained model with an external scorer trained on AI-synthesized datasets. Despite leveraging auxiliary models and additional data sources, our approach outperforms collaborative methods across three out of four datasets. Notably, on ASQP-REST15, our method achieves an improvement of approximately $3\%$ using a single backbone model only without extra synthetic data.

\subsection{Ablation Study}
We conduct an ablation study to evaluate the contribution of each component, the results of which are summarized in Table \ref{tab:ablation study}. Replacing the listwise loss with standard pairwise DPO leads to a decline in F1 score on three datasets, underscoring the advantage of our method in promoting overall structural validity and relational coherence. Especially on ACOS-Rest, where pairwise DPO fails to improve upon the SFT baseline, our approach still achieves a performance gain. The only exception is ASQP-Rest16, where pairwise DPO outperforms the listwise variant. This can be attributed to the fact that listwise DPO leverages overall structural ranking for optimization, using balanced preference data that covers various element-wise confusions. While this avoids the preference fragmentation and sampling bias inherent in pairwise comparisons, it may slightly dilute the focus on specific single-element errors. In contrast, pairwise DPO provides a direct comparison between individual samples, making it particularly effective for correcting single-element mistakes. As detailed in Section \ref{error analysis}, such errors constitute the majority of errors in the ASQP-Rest16 dataset. 

When DPO is removed entirely (i.e., SFT only), performance drops by up to $0.6$ points on ACOS-Laptop. Furthermore, when rationale prediction is also ablated, we observe a sharp decline across all datasets, highlighting the importance of rationale generation in explicitly modeling inter-element relationships and consistently improving both SFT and DPO training.

\begin{figure*}[ht]
  \includegraphics[width=\linewidth]{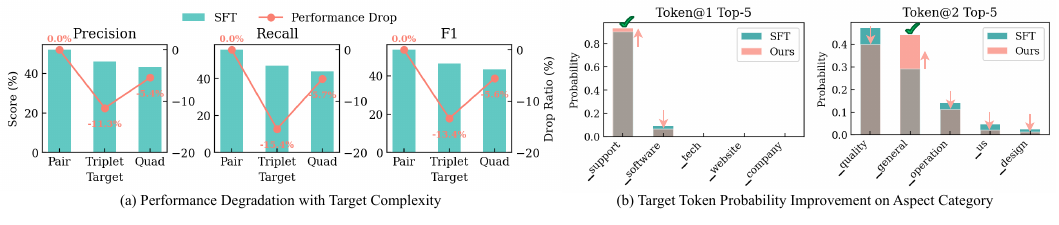}
  \caption {(a) Performance degradation as output structure complexity increases under standard supervised fine-tuning (SFT). A sharp decline is observed when expanding prediction targets from pairs (\textit{a, o}) to triplets (\textit{a, o, c}). (b) Improvement in target token probability for the challenging higher-order (e.g., $c$) element achieved by our method, compared to SFT-only with our reasoning-based template.}
  \label{fig:token shift}
\end{figure*}
\subsection{Impact on Target Token Probability}
To diagnose the performance limitations of SFT-based methods in ASQP, we examine how model performance varies with increasing structural complexity of the output. As illustrated in Figure~\ref{fig:token shift}(a), we observe a notable degradation in performance as the prediction target expands from pairs (e.g., ($a,o$)) to triplets (e.g., ($a,o,c$)) and finally to full quads. The most pronounced decline occurs during the transition to triplets, suggesting that models struggle significantly with inferring \textbf{higher-order elements} such as aspect categories ($c$) and sentiment polarity ($s$). These elements require the model to not only identify individual terms but also to capture \textbf{complex contextual dependencies and relational reasoning} with reference to preceding elements (e.g., linking $a$ and $o$ to deduce c).

Figure~\ref{fig:token shift} (b) further demonstrates the effectiveness of our proposed preference optimization framework. By explicitly modeling inter-element relationships and introducing listwise preference optimization over element-wise confusable candidates, our approach enhances the model ability to disambiguate challenging elements. Specifically, the method increases the target token probability for difficult-to-infer elements such as aspect categories, while suppressing probabilities for incorrect or less relevant tokens. This indicates that the listwise objective encourages more focused and contextually aware predictions, leading to improved relational coherence and final quad accuracy. 

\subsection{Error Analysis}
\label{error analysis}
\begin{figure*}[ht]
 \centering
  \includegraphics[width=0.95\linewidth]{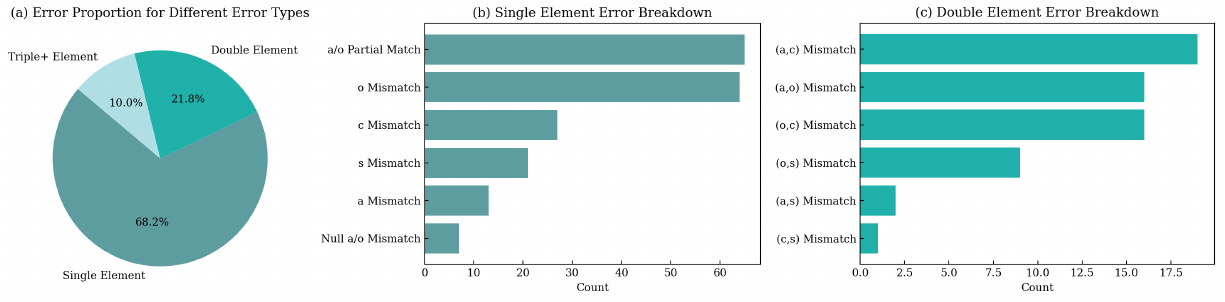}
  \caption{Error analysis on ASQP-Rest16. Figure (a) demonstrates the proportion of coarse-grained error types. Figures (b) and (c) provide the fine-grained error breakdown for the two main error types.}
  \label{fig:error}
\end{figure*}
To further investigate the error scenarios, we conduct a fine-grained error analysis on the ASQP-Rest16 dataset, where listwise preference optimization exhibits limited performance gains compared to pairwise DPO. The results are summarized in Figure \ref{fig:error}.

\paragraph{\textbf{Single-element Errors Dominate.}} In Figure (a), the majority of errors fall into the single-element category, where exactly one element is incorrect while the remaining three are accurately predicted. Within this category, partial matches in aspect ($a$) or opinion ($o$) terms constitute the most common error type, as illustrated in Figure \ref{fig:error} (b). These partial matches often involve predictions that refer to the same underlying concept as the ground truth but exhibit slight discrepancies in syntactic span or expression. 
Table \ref{ao partial match} provides a case study, where the predicted spans are plausible, and sometimes arguably more semantically appropriate than the ground truth. Such inconsistencies frequently \textbf{stem from annotation subjectivity or variability} rather than genuine model inference errors, a phenomenon also observed in prior work \cite{SimRP}. Therefore, our approach, while effective in improving structural validity and relational coherence, may be limited to correcting errors that originate from human annotation bias, particularly when training data are scarce. These observations highlight the need for more consistent, objective, and balanced labeling protocols.

Additionally, opinion ($o$) mismatch constitutes the second most common type of error. In these instances, the model generates opinion expressions that are semantically similar to the ground truth but lexically divergent, while all other elements remain correct. Table \ref{o mismatch} presents an instance where the predicted opinion captures a relationship closer to the surrounding context than the ground truth. Although our listwise preference optimization approach improves the overall ranking of the target quadruple-rationale prediction among confusing candidates, it may reduce emphasis on single-element distinctions compared to pairwise DPO, which explicitly contrasts one positive-negative pair each time. This effect is particularly noticeable in cases where specific element errors dominate the error distribution. 
\begin{table*}[h]
 \centering
  \small
  \caption{An example illustrating the error type of $a$/$o$ \textit{partial match} in ASQP-Rest16, with the quad presented in ($a,o,c,s$) order.}
  \label{ao partial match}
  \begin{tabular}{lp{12cm}}
    \toprule
     \multirow{3}*{{\textbf{Prompt}}} & Given the input text: \{Input Text\}, infer aspect terms, opinion terms, aspect categories, and sentiment polarity following the format. Please join with semicolon if multiple aspects or opinions are detected.\\
     & $\#$Output Format\\
     & (aspect term: [aspect term], opinion term: [opinion term], aspect category: [aspect category], sentiment polarity: [sentiment polarity], rationale: [aspect category] is [sentiment polarity] because [aspect term] is [opinion term]) \\
     \midrule
     \multicolumn{2}{c}{{\textbf{Error Type - \textit{$a$ partial match}}}}\\
     \midrule
      \textbf{Input Text} & Fancy pieces of exotic fish on a \$ 100 dollar plate and not one was eatable.\\
     \midrule
     \textbf{Ground Truth Quad} & (`plate', `100 dollar', `food prices', `bad'); (`\colorbox{pink!40}{exotic fish}', `not one was eatable', `food quality', `bad') \\
     \midrule
     \textbf{Predicted Quad} & (`plate', `100 dollar', `food prices', `bad'); (`\colorbox{pink!40}{fish}', `not one was eatable', `food quality', `bad') \\
     \midrule
     \multicolumn{2}{c}{{\textbf{Error Type - \textit{$o$ partial match}}}}\\
     \midrule
     \textbf{Input Text} & i have been eating at this place for over 8 years now and i have never had one bad meal.\\
     \midrule
     \textbf{Ground Truth Quad} & (`meal', `\colorbox{pink!40}{bad}', `food quality', `great') \\
     \midrule
     \textbf{Predicted Quad} & (`meal', `\colorbox{pink!40}{never had one bad}', `food quality', `great') \\
    \bottomrule
  \end{tabular}
\end{table*}

\begin{table*}[h]
 \centering
 \small
  \caption{An example illustrating the error type of $o$ \textit{mismatch} in ASQP-Rest16, with the quad presented in ($a,o,c,s$) order.}
  \label{o mismatch}
  \begin{tabular}{lp{12cm}}
    \toprule
     \textbf{Input Text} & The service ranges from mediocre to offensive.\\
     \midrule
     \textbf{Ground Truth Quad} & (`service', `\colorbox{pink!40}{mediocre}', `service general', `bad') \\
     \midrule
     \textbf{Prediction} & (aspect term: service, opinion term: offensive, aspect category: service general, sentiment polarity: bad, rationale: service general is bad because service is offensive) \\
     \midrule
     \textbf{Predicted Quad} & (`service', `\colorbox{pink!40}{offensive}', `service general', `bad') \\
    \bottomrule
  \end{tabular}
\end{table*}

\subsection{Influence of Parameters}
\label{sec:parameters}
\begin{figure}[h]
  \includegraphics[width=\linewidth]{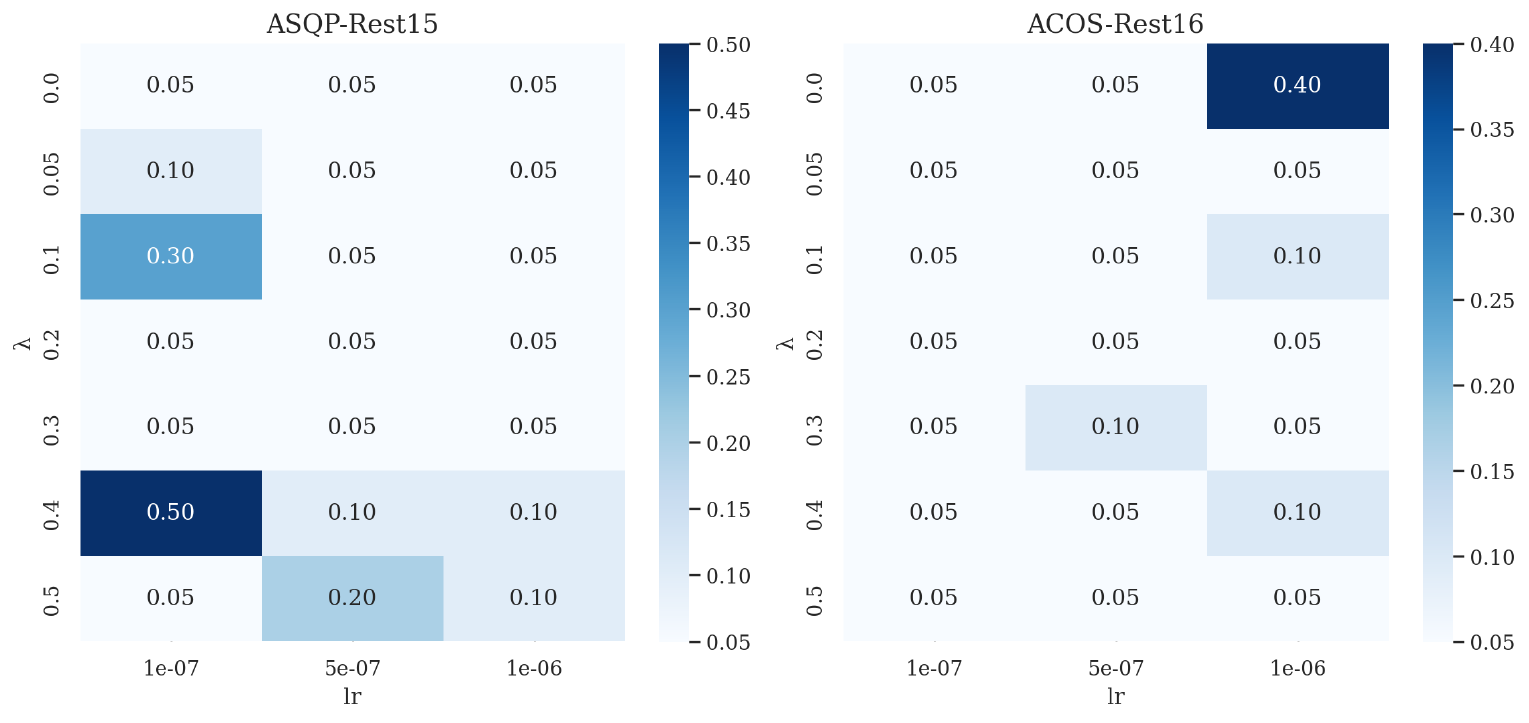}
  \caption{Heatmaps for beta at best F1 score under different learning rate (lr) and $\lambda$ settings on ASQP-Rest15 and ACOS-Rest16 datasets.}
  \label{fig:rest15_heatmap}
\end{figure}

\begin{figure}[h]
  \includegraphics[width=\linewidth]{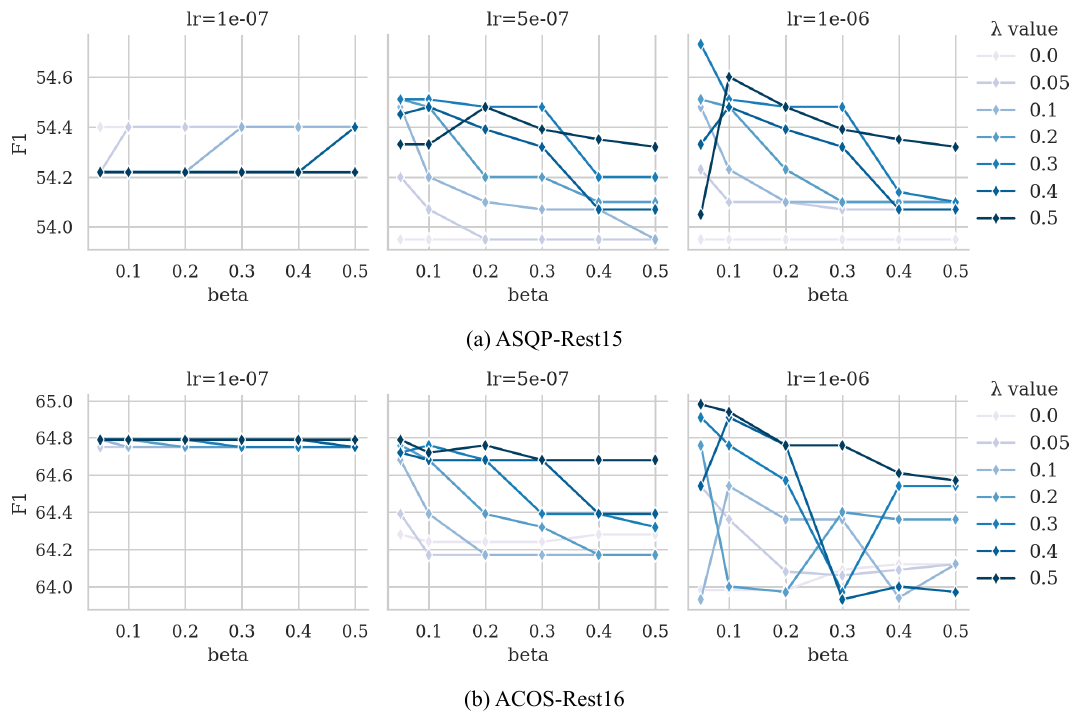}
  \caption{F1 score performance evaluated on ASQP-Rest15 and ACOS-Rest16 under different parameter settings, including beta rate, learning rate (lr), and $\lambda$.}
  \label{fig:rest15_para}
\end{figure}
When applying listwise preference optimization, we employ a hybrid CE loss controlled by hyperparameter $\lambda$ to stabilize the training process. Figure \ref{fig:rest15_heatmap} presents the heatmap for the best beta ratio at the best F1 score under different learning rates and $\lambda$ values on ASQP-Rest15 and ACOS-Rest16 datasets, while Figure \ref{fig:rest15_para} demonstrates the performance trend on these two datasets under different parameter settings. 

It can be observed that:
\begin{itemize}
    \item A beta value of $0.05$ generally yields the best performance across most settings.
    \item A learning rate of $1e-6$ proves more effective for our backbone model in enhancing preference learning.
    \item As $\lambda$ increases, better results are generally achieved with lower values of the beta ratio.
\end{itemize}
This trend can be explained by the roles of these two hyperparameters: $\lambda$ controls the emphasis on the SFT objective within the hybrid CE loss, anchoring the model to SFT knowledge that promotes structural validity and accuracy. On the other hand, the beta ratio $\beta$ modulates the degree of deviation from the reference model. Therefore, when $\lambda$ is large, the training signal is already strongly anchored to SFT. Pairing this with a lower beta further limits deviation from the reference model, which stabilizes optimization and prevents large, potentially noisy shifts driven by listwise preferences, while still allowing incremental improvements. Conversely, when 
$\lambda$ is smaller, a higher beta can be beneficial to permit larger deviations that flexibly capture preference signals. 

\begin{figure}[h]
  \includegraphics[width=\linewidth]{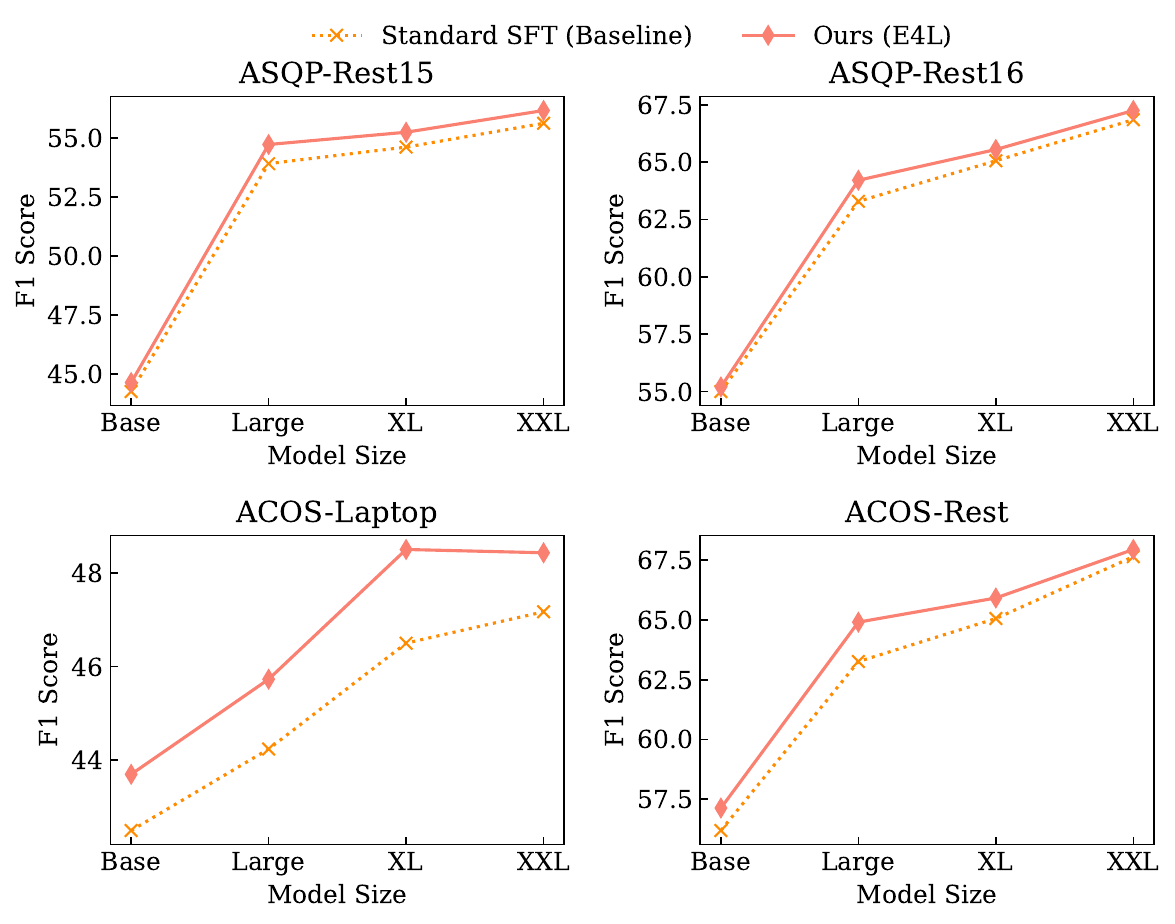}
  \caption{F1 performance when scaling the backbone model Flan-T5 across four datasets. Standard SFT uses instruction prompts without rationale outputs.}
  \label{fig:model size}
\end{figure}

\subsection{Influence of Model Size}
\label{sec:model size}
Figure \ref{fig:model size} reports the F1 score performance as we scale up the backbone model across four datasets. Standard SFT improves monotonically with model size. While our method consistently yields additional gains over SFT by applying listwise preference optimization to confusable candidates that explicitly model inter-element relations, with the largest improvements on ACOS-Laptop. Mid-size models (e.g., large and XXL sizes) benefit the most, as they have sufficient capacity to differentiate confusable alternatives and exploit relational rationales without saturating. Conversely, small models show modest gains because limited capacity constrains their ability to leverage richer rationales and preference signals. For the largest model, incremental gains seem to diminish. This ceiling effect is expected because (i) SFT alone already achieves strong performance, leaving limited headroom, and (ii) a fixed candidate set may under-challenge very large models, reducing the marginal value of preference signals. 


\section{Related Work}

\paragraph{Aspect Sentiment Quad Prediction.} ASQP is widely regarded as the most challenging task in ABSA, as it requires the simultaneous prediction of four sentiment elements \cite{tkde/ZhangLDBL23, lai2025llmsteamup, naacl/ZhangDLPB24}. Prior efforts have largely focused on data augmentation techniques to improve performance \cite{coling/GaoFLLLLBY22, starsem/WangJMLO23, acl/MvP, acl/bsvp, self-scorer}. With LLMs demonstrating strong generalization and reasoning via in-context learning (ICL) \cite{Brown2020LanguageMA, Kojima_Shixiang_Gu_Reid_Matsuo_Iwasawa, Wei2021FinetunedLM, nips/AnML0LC24} and chain-of-thought (CoT) prompting \cite{Wei2022ChainOT, Zhang_Zhang_Li_Smola_2022, Wei_Wang_Schuurmans_Bosma_Chi_Le_Zhou, Yao2023TreeOT}, recent research has shifted toward leveraging LLMs to improve reasoning in ASQP. For example, SimRP \cite{SimRP} leveraged syntactic and semantic retrieval to improve demonstration selection for ICL. While Kim et al. \cite{self-consistency} introduced self-consistency mechanisms to help models internalize the rationales behind predictions from LLMs. Nevertheless, LLMs still exhibit a noticeable performance gap in accurately predicting sentiment quads \cite{naacl/ZhangDLPB24, lai2025star}, underscoring the need to enhance structured reasoning. 

\paragraph{Preference Optimization for LLMs.} To improve LLM performance and reliability, Reinforcement Learning from Human Feedback (RLHF) \cite{nips/ChristianoLBMLA17, Ouyang0JAWMZASR22} is an effective approach to align models using human preference data, but it typically requires complex reward modeling and substantial computational overhead. Direct Preference Optimization (DPO) \cite{DPO} simplifies this by optimizing directly on pairwise preferences, and has been widely adopted for complex reasoning, particularly in mathematics \cite{stepdpo, yang2025supercorrect}. While the base models for optimization typically already possess a strong reasoning capacity. In affective computing, preference optimization has primarily been applied to generation tasks like conversational generation \cite{wang2023aligning, decoupledESC}, but it is rarely explored for structured sentiment analysis such as ASQP. 


\section{Conclusion}

This work addresses challenging ASQP by introducing a listwise preference optimization framework that leverages element-wise confusable candidates to enhance structural validity and relational coherence.
Prior methods largely relied on marker-based prediction with SFT, which lacked explicit modeling of inter-element relationships and struggled to distinguish closely related candidates. To overcome these limitations, our approach employs a reasoning-based natural language template that jointly generates sentiment quadruples and rationales under element prefixes. Then, element-wise confusable candidates are constructed through syntactic and semantic proximity, where one or more elements within the template are altered to challenge model understanding and reasoning. Beyond SFT and conventional DPO, our model is aligned with preference optimization using a listwise objective, which places the gold output above all competing candidates. Extensive experiments across four benchmark datasets demonstrate that our approach consistently outperforms baselines, achieving improved performance in quad prediction accuracy and explanation consistency.



\bibliographystyle{ACM-Reference-Format}
\bibliography{main}

\appendix

\end{document}